\pdfoutput=1

\documentclass[11pt]{article}

\usepackage[final]{acl}

\usepackage{times}
\usepackage{latexsym}

\usepackage[T1]{fontenc}

\usepackage[utf8]{inputenc}

\usepackage{microtype}

\usepackage{inconsolata}
\usepackage{booktabs}

\usepackage{graphicx}
\usepackage{hyperref}
\usepackage{url}
\usepackage{appendix}
\usepackage{subcaption}

\title{Are Generative Language Models Multicultural? A Study on Hausa Culture and Emotions using ChatGPT}

\author{Ibrahim Said Ahmad, Shiran Dudy, Resmi Ramachandranpillai \and Kenneth Church \\
         Northeastern Univeristy \\ Boston, MA, USA \\
         i.ahmad@northeastern.edu}

\begin{document}
\maketitle
\begin{abstract}
Large Language Models (LLMs), such as ChatGPT, are widely used to generate content for various purposes and audiences. However, these models may not reflect the cultural and emotional diversity of their users, especially for low-resource languages. In this paper, we investigate how ChatGPT represents Hausa's culture and emotions. We compare responses generated by ChatGPT with those provided by native Hausa speakers on 37 culturally relevant questions. We conducted experiments using emotion analysis and applied two similarity metrics to measure the alignment between human and ChatGPT responses. We also collected human participants ratings and feedback on ChatGPT responses. Our results show that ChatGPT has some level of similarity to human responses, but also exhibits some gaps and biases in its knowledge and awareness of the Hausa culture and emotions. We discuss the implications and limitations of our methodology and analysis and suggest ways to improve the performance and evaluation of LLMs for low-resource languages.
\end{abstract}

\section{Introduction}
Large Language Models (LLMs), such as ChatGPT are rapidly becoming popular and are employed in generating content for various purposes, be it for personal notes, in the workplace, or even for research and education. Additionally, these models have a global reach, meaning a diverse set of people with varying cultural backgrounds. As a result, these models need to reflect the cultural differences of people and their emotional sensitivities when generating content. Since these models were trained mainly on Internet data, there is a high probability that they will be biased toward cultures with languages that are highly resourceful, such as English, Japanese, Chinese, German, and French \cite{arora-etal-2023-probing, lucy-bamman-2021-gender, bias-out2021}. 

Previous studies have shown that LLMs exhibit intersectional biases \cite{bias-out2021}, gender stereotypes \cite{lucy-bamman-2021-gender}, and political biases \cite{rozado2023political}. In this paper, we study cultural differences surrounding the representation of Hausa culture and emotions in ChatGPT. The relationship between language, culture, and emotions has been well established in the literature \cite{russell_culture_1991,Anna1992,andrew2021}. 

Language serves as a medium through which cultural identities, values, and traditions are expressed and transmitted. Cultural narratives, metaphors, and discourses are embedded in language, reflecting the cultural heritage of the community \cite{language-culture-2014}. The language we use can influence how we experience and express emotions. Language can both dampen and intensify emotional experiences. Journaling or verbal expression of emotions, for example, can help regulate negative emotions \cite{emotion-culture}. 

Our work focuses on the extent to which multilingual LLMs generate culture-aware responses. We aim to investigate the validity of cultural and emotional responses generated by multilingual LLMs for low resource languages. In our experiments, validity is assessed by speakers of the Hausa language. Despite being spoken by approximately 100 to 150 million people globally, Hausa remains a low-resource language. Hausa is spoken mostly in West African countries such as Nigeria, Nijer, Ghana, Cameroon and Benin \cite{pawlak-Nina-2023}. 

The remainder of this paper is organized as follows: Section 2 details the experiment design, Section 3 provides the results, Section 4 discusses the findings and finally, the conclusion is described in Section 5.



\section{Experiments}
As a first step, we prompt ChatGPT with 37 questions that are expected to yield culturally-aware responses (more details below). Then, we used those as survey questionnaire, and collected the responses from native Hausa speakers living in Nigeria. We collected two types of responses; first, we asked participants to answer the questions as open-ended questions. Then, we asked them to rate the responses generated by ChatGPT using a 5-point Likert scale. Our experiments involve a comparative analysis of the survey responses and those generated by ChatGPT.

\subsection{Data}
We used a total of thirty-seven (37) prompts (or questions) that are expected to produce responses that are culturally dependent. Eighteen (18) are from a prior study by \citet{havaldar-etal-2023-multilingual}. The remaining Nineteen (19) questions were crafted using literature on African cultures and emotions. The prompts were validated by a psycholinguistic expert in the Hausa language and culture.

We prompt ChapGPT using a similar technique in \citet{havaldar-etal-2023-multilingual}, where each question is preceded by a fixed pre-question prompt: \textit{"You are a helpful chatbot. Your goal is to answer my questions like you are a human capable of feelings and emotions. You live in Northern Nigeria. Answer the following question using a single sentence that begins with 'I would feel...'"}. 

We engaged 18 individuals who are native speakers of the Hausa language and identify as having Hausa ethnic background, to (1) evaluate the cultural alignment of the responses generated by ChatGPT and (2) collect their (human) responses to the same questions (or prompts) using a survey questionnaire. The questionnaire is divided into two sections as follows: 

\begin{enumerate}
    \item \textbf{Open-ended questions:} Participants were prompted to provide answers to the thirty seven (37) questions, with a requirement of at least five words per response. 
    \item \textbf{Psychometric scale:} Participants were presented with both the question posed and the response generated by ChatGPT. They were then instructed to assess the extent to which they agreed with the cultural representation and emotional content conveyed in ChatGPT responses. Table \ref{tab:prompts_label} shows an example of the prompts and the corresponding responses by ChatGPT. 
\end{enumerate}

\begin{table*}[t]
    \centering
    \caption{Examples of prompts and the corresponding responses from ChatGPT  }
    \begin{tabular}{p{0.3cm} p{6cm} p{6cm}}
        \hline \hline
        \textbf{SN} & \textbf{Prompt}   & \textbf{Response} \\
        \hline
        1. & How would you feel if your customer asked for a discount on your product?  & I would feel open to discussing the request for a discount, aiming to find a fair solution that values the customer's needs and the product's worth. \\
        2. & How would you feel about prioritizing eating at a restaurant over cooking at home? & I would feel excited about the prospect of enjoying a meal at a restaurant, appreciating the convenience and the opportunity for a shared experience. \\
        3. & How would you feel if your son-in-law greets you with a handshake? & I would feel a bit surprised but understanding, recognizing that cultural norms and personal preferences can differ in greetings. \\
        4. & How would you feel about moving your aged parents to a nursing home? & I would feel a mix of emotions, including concern for their well-being and the practical considerations, while also grappling with the emotional aspect of making such a decision for my aging parents. \\
        5. & How would you feel if you were served food without a spoon? & I would feel a bit puzzled but adaptable, considering alternative ways to enjoy the meal without a spoon. \\
        \hline
    \end{tabular}
    \label{tab:prompts_label}
\end{table*}

\subsection{Analysis}
To assess the alignment between human textual responses and those generated by ChatGPT, we performed 
emotion analysis per each of the textual sources. We also applied two similarity metrics, BERTScore \cite{zhang2019bertscore} and METEOR (Metric for Evaluation of Translation with Explicit Ordering) \cite{banerjee2005meteor} to compare the textual similarities of the responses. 

\textbf{Emotion Analysis} is employed to characterize the emotional distribution of each textual source in order to compare and assess the \textit{emotional spectrum} between humans and ChatGPT. The emotion labels to classify responses were: `positive', `negative', `compoud', and `neutral'.

\textbf{BERTScore} computes a \textit{semantic similarity} score for each token in the generated sentence with each token in the reference sentence and the token similarity is computed using contextual embeddings \cite{zhang2019bertscore}. \textbf{METEOR} is a metric employed for machine translation evaluation based on unigram matching between machine translation and human reference translations. The score is computed using a combination of unigram precision, unigram recall, and a fragmentation measure that is designed to directly capture how well the words matched in machine translation are arranged in relation to the reference \cite{banerjee2005meteor}. METEOR was employed by \citet{bhatt2024extrinsic} for open-ended questions. To this end, METEOR, would help assess the usage of specific word choices complementary to BertScore measuring semantic similarity.


\section{Results}
\subsection{Emotional Analysis of Participants' Responses and ChatGPT Responses}
\begin{figure*}[t]
    \centering
    \begin{subfigure}{0.45\textwidth}
        \includegraphics[width=\linewidth]{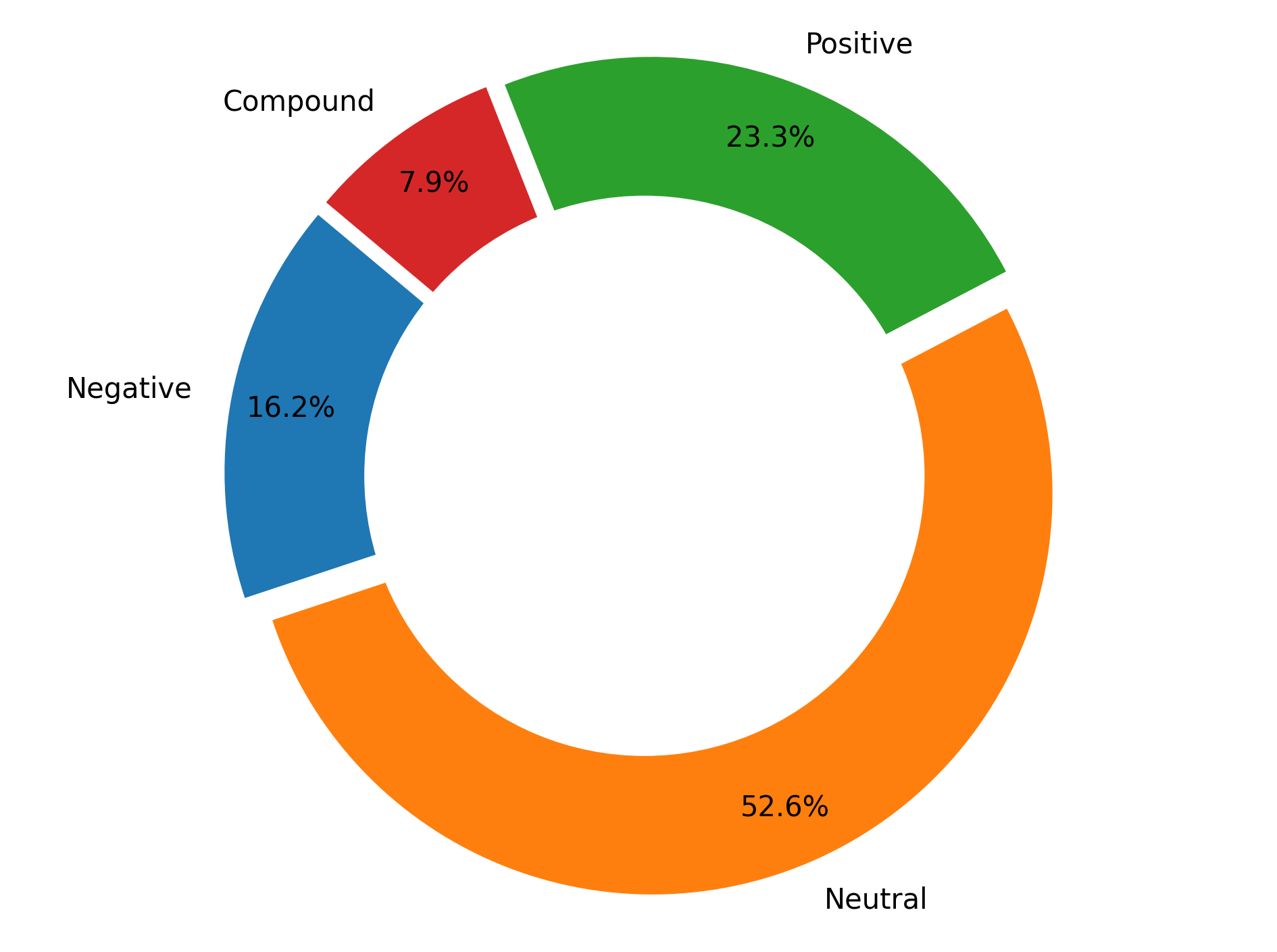}
        \caption{Participants emotional classes. The responses contains `positive', `negative', and `compound', and `neutral' classes.}
        \label{fig:sub1}
    \end{subfigure}
    \hfill
    \begin{subfigure}{0.45\textwidth}
        \includegraphics[width=\linewidth]{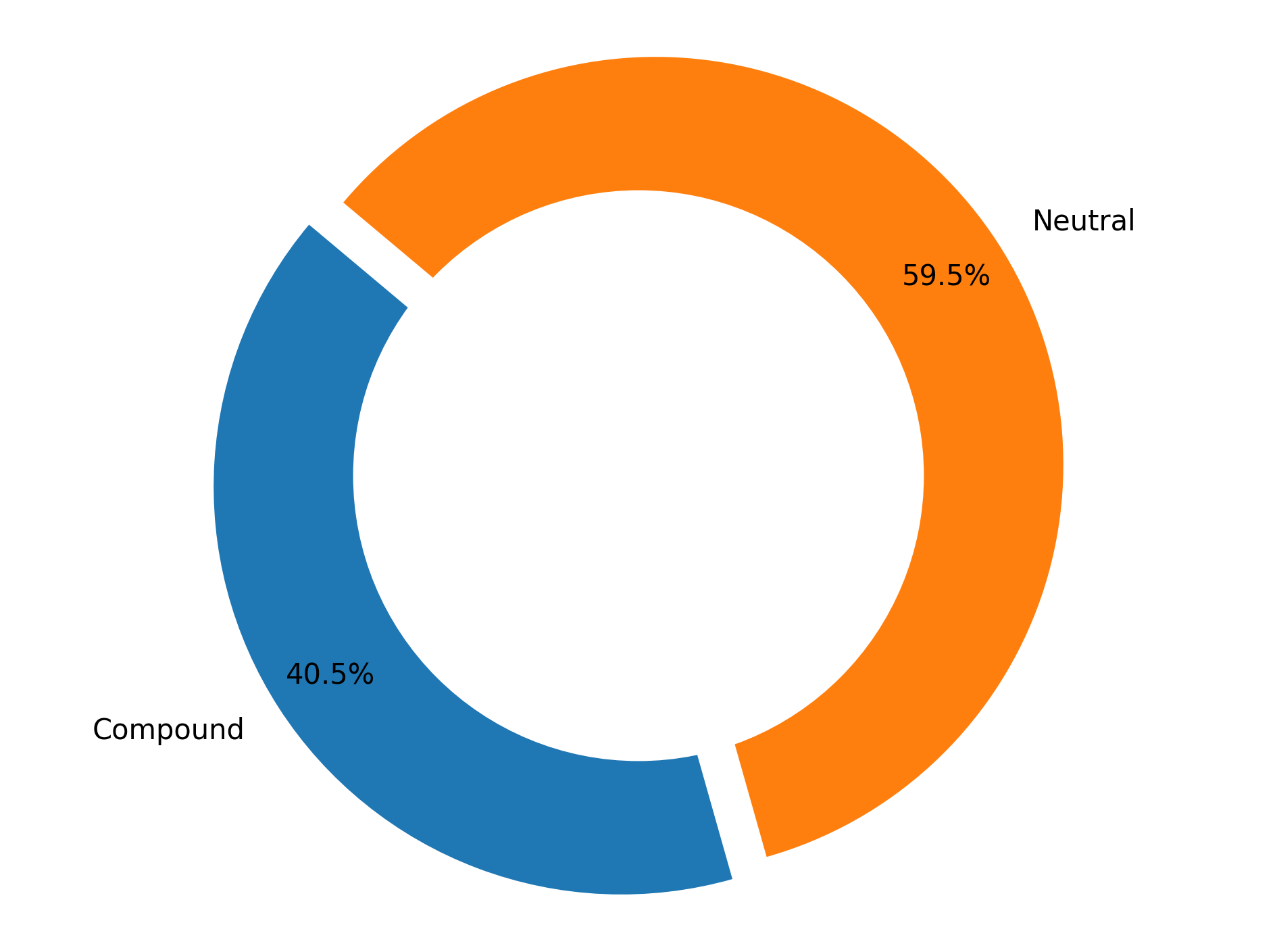}
        \caption{ChatGPT emotional classes. The responses are emotionally `compound' and `neutral'.}
        \label{fig:sub2}
    \end{subfigure}
    \caption{Emotion analysis for Participants and ChatGPT responses.}
    \label{fig:emotional}
\end{figure*}

Figure \ref{fig:emotional} illustrates the results of the emotion analysis. 
The emotion analysis reveals that ChatGPT-generated responses are limited in their emotional expression; a majority is neutral, and the rest is compound. The participants' generated responses, on the other hand, offer a richer emotional spectrum exhibiting also `positive' and `negative' sentiments. This finding suggests an emotional mismatch where ChatGPT produces more plain neutral text relative to the human nature of the participants in our study.  


\subsection{Textual Similarities of Participants' Responses and ChatGPT's Responses}
\begin{figure*}[t]
    \centering
    \includegraphics[width=\textwidth]{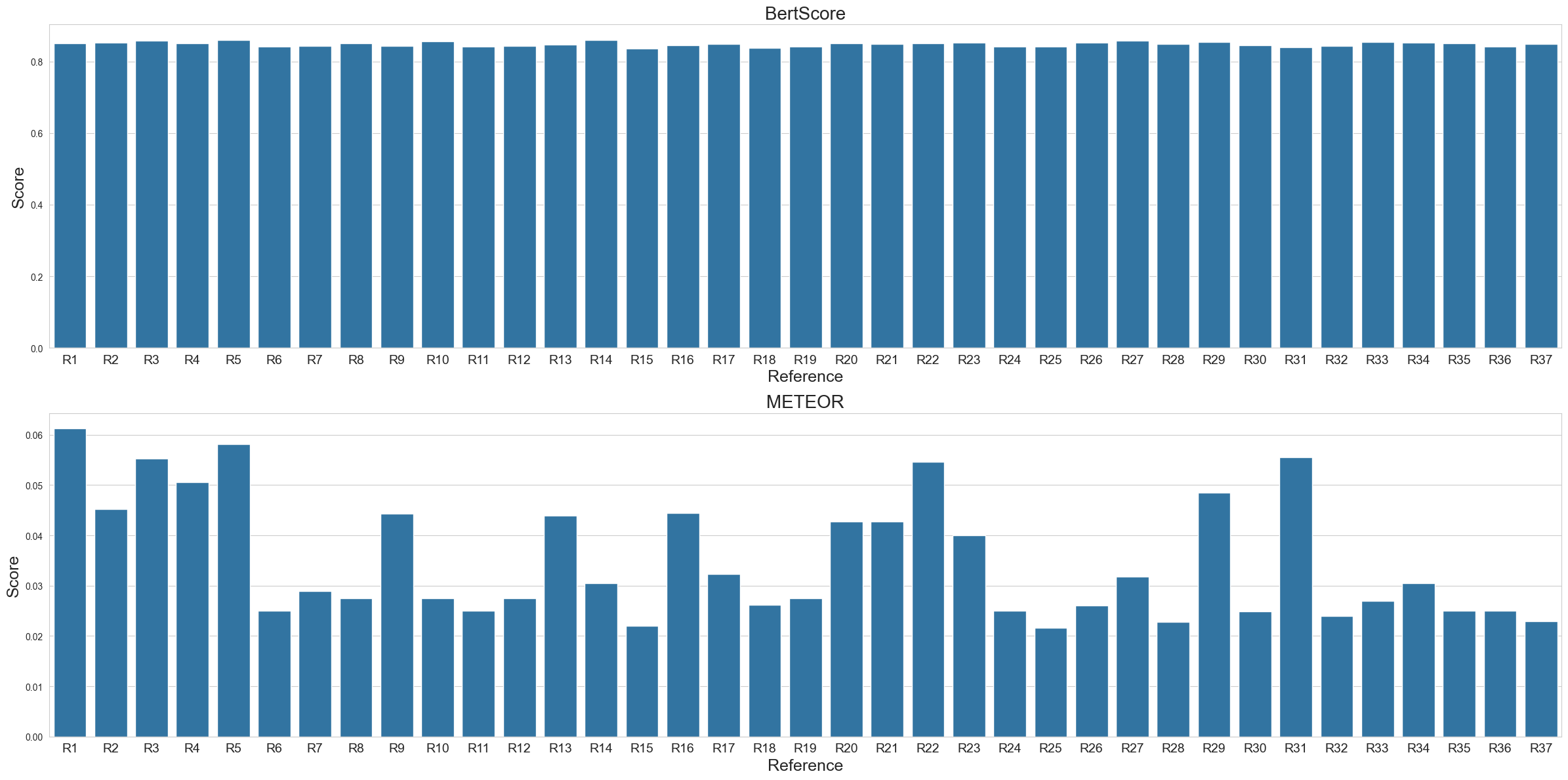}
    \caption{Median similarity scores between responses returned by ChatGPT and Human Responses Recorded. There is a single response for each prompt per ChatGPT and 18 human responses. Each ChatGPT response is compared to the human responses and the median similarity scores were recorded for the 37 prompts.}
    \label{fig:similarity}
\end{figure*}

The comparison between the responses of the participants and those produced by ChatGPT using BERTScore and METEOR is presented in Figure \ref{fig:similarity}. Assessing the textual similarity between responses generated by LLMs and those created manually remains an evolving field of research. Consequently, there are currently no flawless metrics available for this purpose. 

The result of the textual similarity using the BertScore and METEOR may shed a different light on the same story. Although BertScore shows very strong semantic similarities between participants to CHatGPT, METEOR shows relatively lower similarities between the two textual sources. This could be explained based on the different architectures of the metrics. While BertScore tends to focus on capturing the overall semantic similarities, METEOR considers specific linguistic aspects such as word overlap, stemmed tokens, and synonymy. METEOR might assign lower scores if the generated text deviates from the reference in terms of surface-level features, even if they convey similar meanings. Therefore, at this point we can conclude that while responses may be semantically similar to participants` ones, it is unclear whether the wording, and word choices is appropriate to reflect cultural characteristics. In order to further learn about the authenticity of responses, in the next step we asked participants to directly score `how well the ChatGPT responses sound like a native Hausa speaker.

\subsection{Humans Assessment for ChatGPT Cultural Alignment with Hausa Culture}
\begin{figure*}[t]
    \centering
    \includegraphics{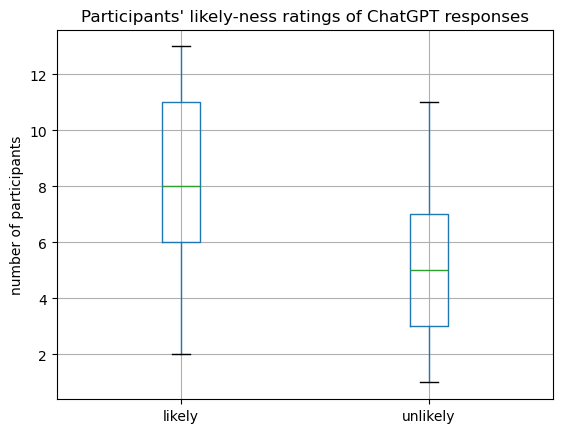}
    \caption{Participants likely-ness rating of ChatGPT responses. While there are 8.2 subjects on average who find ChatGPT responses to be likely to uttered by native speakers of Hausa, there are 5.3 who find these responses unlikely (The plot indicates median, the average was computed separately).}
    \label{fig:boxplot}
\end{figure*}

18 Participants were instructed to use the Likert scale to assess 37 ChatGPT responses, and particularly to indicate the degree to which these responses reflect the culture and emotions of the Hausa people. Participants were asked to rate each response on a scale of 1 to 5, with 1 indicating that the response is not likely to be uttered (by a native speaker) and 5 meaning it is likely to what they would expect. In order to process the results, we follow the three steps. First, we merged the rating scores 1 and 2 to mean that the response is \textit{unlikely}, 3 to mean undecided, and 4 and 5 to mean that the response is \textit{likely}. Second, per each question we counted how many people (of the 18 participants) rated \textit{likely}, and how many rate \textit{unlikely} across all 37 ChatGPT responses, third, we plotted the corresponding boxplot in Figure \ref{fig:boxplot} and computed the average number of people who found it likely and unlikely respectively. Our finding suggests that on average 8.2 people find ChatGPT's responses likely, while we have 5.2 people on average who do not see these responses as likely to be spoken by a native speaker. This finding suggests that even though responses may be semantically similar, there remains a cultural mismatch rendering responses to be `not quite there' with regards to the range of plausible anticipated responses.




\section{Discussion}



The limited cultural alignment found in this work can be attributed to several factors such as the quality and quantity of the training data, the fine-tuning process, the cultural and linguistic diversity of the users, and the evaluation methodology. 
These factors are further discussed below.

\textbf{Training data:} Humans grow up learning about cultural values in their lives. This shapes how they understand and respond to cultural questions. However, LLMs lack this background knowledge or life experience \cite{herbold2023large, lu2023error, steiss2024comparing, gao2023comparing}. So, when it gathers and organizes information, it doesn't understand the context or importance of cultural diversity. This is due to the fact that ChatGPT is trained on a large corpus of conversational data, which may not reflect the cultural and emotional diversity of its users, especially users of low-resource languages like Hausa. The training data may also contain noise, errors, or biases that may affect the model's performance. For example, the training data may be skewed towards certain domains, topics, or perspectives that may not be relevant or appropriate for Hausa speakers \cite{bender2021}. Therefore, the quality and quantity of the training data are crucial for developing more culture-aware and emotion-sensitive LLMs for low-resource languages \cite{bhat2023large}.

\textbf{Fine-tuning process:} ChatGPT is fine-tuned on a smaller dataset of multilingual conversations, which may not capture the nuances and variations of Hausa culture and emotions \cite{qiu2024chain}. For example, to the prompt~\footnote{We note that each question like the following has additional text instructing it to answer as a Hausa speaker} \textit{"How would you feel if your student call you by your first name"}, ChatGPT's response was \textit{"I would feel a bit surprised but might appreciate the informality, recognizing that the student may see our relationship as more casual."}.  
This is an example for an inappropriate response in the Hausa culture and traditions and it will be considered very disrespectable to call your teacher by their first name even in an informal setting. Therefore, the fine-tuning process needs to be carefully designed and optimized for developing more robust and adaptable LLMs for low-resource languages.

Another example, for the question: \textit{"How will you feel about prioritizing eating in a restaurant over cooking at home?"}, more than 68\% of the participants considered ChatGPT's response as culturally inappropriate. Similarly, for the question: \textit{"How would you feel about your ward moving out of the house at the age of 18?"},  less than 2\% of the respondents agreed with the response.


\textbf{Cultural and linguistic diversity:} ChatGPT is designed to generate responses for a diverse set of users, who may have different cultural backgrounds, values, and preferences. However, when evaluating on Hausa, Chatgpt did not seem to capture these cultural and linguistic diversities indicated in lower METEOR scores. In particular, ChatGPT may not be able to produce the Hausa dialects, idioms, or expressions that may exist within the Hausa language and culture. ChatGPT may also not be able to adapt to the different contexts, situations, or goals that may influence the model's performance. Therefore, the cultural and linguistic diversity of the users poses a challenge and an opportunity for developing more personalized and context-aware LLMs for low-resource languages.


Based on these factors, we suggest some ways to improve the performance and evaluation of ChatGPT and LLMs for low-resource languages, such as Hausa. First, we suggest using more diverse and representative data that can cover more topics and scenarios that Hausa speakers may encounter in the digital world. For example, we can use data from different sources, such as social media, news, blogs, or forums. We can also use data from different groups of people, such as age, gender, education, or location. These data can enrich the model's knowledge and adaptability and provide a more realistic and authentic evaluation of the model's performance. 

Second, we suggest incorporating human feedback and perspective that can improve the model's performance. For example, we can use methods such as user testing, surveys, interviews, or focus groups. We can also use techniques such as active learning, reinforcement learning, or dialogue management. These methods and techniques can enhance the learning and interaction of the model and provide a more user-centric and user-friendly evaluation of the model's performance. 

Third, we suggest proposing evaluation metrics that can measure various aspects of natural language and human communication considering coherence, relevance, fluency, or sentiment. These metrics can complement the similarity metrics and provide a more comprehensive and holistic assessment of the model's performance.

\section{Conclusion}
We investigated how ChatGPT, a generative Large Language Model (LLM), represents the Hausa culture and emotions, a low-resource language spoken by over 100 million people in West Africa. We compared the responses generated by ChatGPT with those produced by native Hausa speakers on 37 culturally relevant questions. We employed emotional analysis, semantic and ngram textual analyses. We also collected the ratings and feedback of human participants on the ChatGPT responses, and evaluated their cultural authenticity.

Our results show that ChatGPT has a limited degree of alignment with human responses. We found a mismatch in the emotional diversity exhibited in the responses of the participants compared to the responses of ChatGPT. We showed that while artificial responses were semantically similar to human participants, they were not aligned with anticipated word choices. Finally, participants found that some ChatGPT responses were likely, but that others were unlikely to be spoken by a member of their culture.  


Our study highlights the imperative for improving ChatGPT and other LLMs' performance and evaluation in low-resource languages to better represent users' cultural and emotional diversity, crucial for sensitive domains like health and education in order to promote equitable and inclusive participation. We suggest utilizing more diverse data, human feedback, and alternative evaluation metrics. In conclusion, our research underscores the importance of evaluating LLMs for low-resource languages, exemplified by Hausa. Future directions involve expanding datasets and establishing crowd-truth~\cite{aroyo2013crowd} approaches to aid validation strategies of cultural alignment evaluations for researchers and practitioners.

\section{Limitations}

Our experiment and findings have limitations. In future experiments, we will consider increasing the robustness of our results by increasing the number of human participants, and by ensuring their demographics is representative of the Hausa population. In addition, for each question, we may benefit comparing our 18 participants responses, to a distribution of responses by ChatGPT, as currently we generated a single ChatGPT response per question. We may also consider additional approaches for word choice or word overlap evaluation such as tf-idf and word-edit distance~\citep{bhatt2024extrinsic} to strengthen the analysis.

We also note that employing sentiment analysis, BertScore evaluation metrics introduce limitations in our work.
Since the Hausa dialect of English has not been trained on the sentiment analysis classifier, the labels may not reflect the realistic emotional labels found in the text. However, since employing this classifier is consistent across both textual sources, the value of using this tool is comparing the label distributions indicating differences across these textual resources.
In addition, the strength BertScore results that compare the semantics of these textual resources is limited as BertScore was not trained on this Hausa variation of English - and may not capture well similarities or differences resulted by synonyms or phrases that are language and dialect specific.

Finally, despite evidence in the literature \cite{bhatt2024extrinsic}, evaluating \textit{open-ended} text that presents great variability, with machine translation tools, where the machine-translated sentence and the human-translated sentence are expected to present similarity to some degree, poses a limitation to our work as well.

\bibliography{custom}

\clearpage
\onecolumn

\appendix
\section{Appendix}\label{sec:appendix}
\begin{table}[h!]
    \centering
    \caption{The questions used as prompts. Note that questions 20 to 37 are adopted from \citet{havaldar-etal-2023-multilingual}. Also we note that each question has additional text instructing ChatGPT to answer as a Hausa speaker.}
    \begin{tabular}{p{0.3cm} p{14cm}}
        \toprule\toprule
        \textbf{SN} & \textbf{Prompt}   \\
        \midrule

        1	&	How will you feel if your customer asks for discount to your product? \\
        2	&	How will you feel about prioritising eating at a restaurant over cooking at home? \\
        3	&	How will you feel if your son in law greets you with a handshake? \\
        4	&	How will you feel about moving your aged parents to a nursing home? \\
        5	&	How will you feel if you are served food without a spoon? \\
        6	&	How would you feel if your student call you by your first name? \\
        7	&	How would you feel when your child say you should get out of his/her room? \\
        8	&	How would you feel if your relative show up in your home unannounced? \\
        9	&	How would you feel if your friend visits you without an appointment? \\
        10	&	How would you feel if your child say you should shut up? \\
        11	&	How would you feel if you go out for a drink and your friend expects you to pay for their drink? \\
        12	&	How would you feel if your friend hugs you in public? \\
        13	&	How would you do while eating at home and you receive a guest? \\
        14	&	How would you feel about moving out of your parents’ house at the age of 18? \\
        15	&	How would you feel if your dad dates your ex-girl friend? \\
        16	&	How would you feel when your daughter dates your ex-boyfriend? \\
        17	&	How would you feel about your ward moving out of the house at the age of 18? \\
        18	&	How would you feel if your lover kisses you in public? \\
        19	&	How would you feel when your child takes you to elderly  care/retirement home? \\
        20	&	How would you feel about turning down the food offered by your neighbour? \\
        21	&	How would you feel about turning down the food offered by your sibling? \\
        22	&	How would you feel when providing critical feedback to your colleague? \\
        23	&	How would you feel if you receive critical feedback from your colleague? \\
        24	&	How would you feel about confronting a colleague in the office? \\
        25	&	How would you feel about confronting your friend in their home? \\
        26	&	How would you feel about confronting your friend in your home? \\
        27	&	How would you feel about prioritising your friends over family? \\
        28	&	How would you feel about prioritising your family over friends? \\
        29	&	How would you feel if your parents moved in with you right after you were married? \\
        30	&	How would you feel if your parents moved in with you when they were very elderly? \\
        31	&	How would you feel about taking off your shoes before entering your friend’s home? \\
        32	&	How would you feel about your friend insisting you take off your shoes before entering their home? \\
        33	&	How would you feel if your guests chose to keep their shoes on when entering your home? \\
        34	&	How would you feel when interacting with the boss of your supervisor? \\
        35	&	How would you feel if you are asked to interact with the boss of your supervisor? \\
        36	&	How would you feel about sharing your excellent performance on a class test? \\
        37	&	How would you feel about sharing your terrible performance on a class test? \\
        
\bottomrule
\end{tabular}
\label{tab:prompts_label_big}
\end{table}

\twocolumn

\end{document}